\documentclass[final,3p,twocolumn]{elsarticle}

\usepackage{hyperref}

\journal{Robotics and Autonomous System}

\usepackage{amssymb}
\usepackage{amsthm}
\usepackage{amsmath}
\usepackage{amsmath}
\usepackage{subfigure}
\usepackage{graphics}
\usepackage{hyperref}
\usepackage{breqn}
\usepackage{titlesec}
\usepackage{xcolor}

\newtheorem{theorem}{Theorem}
\newtheorem{remark}{Remark}
\newtheorem{assumption}{Assumption}
\newtheorem{definition}{Definition}
\newtheorem{property}{Property}

\begin{document}

\begin{frontmatter}

\title{Finite-time disturbance reconstruction and robust fractional-order controller design for hybrid port-Hamiltonian dynamics of biped robots}

\author[mainaddress]{Yousef Farid}
\ead{yousef.farid@unina.it}
\author[mainaddress]{Fabio Ruggiero}
\ead{fabio.ruggiero@unina.it}
\address[mainaddress]{PRISMA Lab, Department of Electrical Engineering and Information Technology, University of Naples Federico II, Via Claudio 21, 80125, Naples, Italy}




\begin{abstract}
In this paper, disturbance reconstruction and robust trajectory tracking control of biped robots with hybrid dynamics in the port-Hamiltonian form is investigated. A new type of Hamiltonian function is introduced, which ensures the finite-time stability of the closed-loop system. The proposed control system consists of two loops: an inner and an outer loop. A fractional proportional-integral-derivative filter is used to achieve finite-time convergence for position tracking errors at the outer loop. A fractional-order sliding mode controller acts as a centralized controller at the inner-loop, ensuring the finite-time stability of the velocity tracking error. In this loop, the undesired effects of unknown external disturbance and parameter uncertainties are compensated using estimators. Two disturbance estimators are envisioned. The former is designed using fractional calculus. The latter is an adaptive estimator, and it is constructed using the general dynamic of biped robots. Stability analysis shows that the closed-loop system is finite-time stable in both contact-less and impact phases. Simulation studies on two types of biped robots (i.e., two-link walker and RABBIT biped robot) demonstrate the proposed controller's tracking performance and disturbance rejection capability.
\end{abstract}

\begin{keyword}
\texttt Bipedal robots \sep Hybrid systems\sep Port-Hamiltonian dynamics\sep Fractional sliding surface\sep Finite-time control\sep Disturbance estimator
\end{keyword}

\end{frontmatter}

\section{Introduction}
In the near future, bipedal robots are expected to be employed in a broad set of applications within industries, service works, and medical activities~\cite{Bruno2009,Chevallereau2009,Hawley2019,Yamamoto2016,Kuindersma2016}. To achieve this goal, it is worth studying humans' outstanding features in-depth, such as robustness, dexterity, and adaptability in different environmental conditions. In the last decade, numerous control algorithms were proposed for the motion control and the stabilization of biped robots \cite{Yeon2014,Hereid2018,Noot2019,Shin2015,Saputra2019,Saputra2016,Righetti2020,Pi2020}. Due to the intrinsic structure complexity, the nonlinearity problems, the existence of discrete dynamics caused by environmental impacts, and the undesired effects of external disturbances, the biped robots' dynamic control is still challenging.

In this paper, the design of a robust controller is carried out to estimate the unknown external disturbances adaptively and, thus, preserve the stability of biped robots against interaction impacts and environmental constraints.

\subsection{Related works}
A hybrid system can describe the dynamic behavior of a typical biped robot having multi-phase properties given by the swing and stance phases of the legs. The impulse effect is caused by the hitting of legs to the walking surface and environmental constraints. Many research works about the control of biped robots in a hybrid form were presented in the literature~\cite{Jessy2014,Hamed2017,Hamed2019,Hamed2012,Gritli2017,Hamed2017_2,Sam2016,Kolathaya2017}. Several works focused on the stability analysis of periodic orbits in walking biped robots using Poincaré maps \cite{Hassane2015,Veer2019,Wang2011,Znegui2020,Yazdi2018}. The main drawback of employing a Poincaré map is that, in practical applications, it must be approximated via numerical methods, and it has no closed-form solutions. 
A hybrid feedback control scheme was proposed in~\cite{Hamed2014} to stabilize the biped robot’s walking and in which the parameters of the central controller were regulated via an event-based technique. The disadvantage of using this method is that there is a considerable delay between the affecting time of the external disturbance and the applying time of the control signal.
On the other hand, many results were presented on the control and stabilization of hybrid dynamical systems. However, several issues still need to be discussed, such as selecting the appropriate Lyapunov functions for stability analysis, investigating finite-time convergence problems, and designing disturbance estimators for hybrid nonlinear systems.

As well known, the dynamic equations of most of the electro-mechanical systems such as electric vehicles, unmanned aerial vehicles, and legged robots can be written in the so-called port-controlled Hamiltonian (pH) form \cite{Duindam2005,Duindam2009,Pei2017,Rashad2019}. In a mechanical pH system, the Hamiltonian function is the sum of the kinetic and potential energies. Besides, it can be labeled as a Lyapunov function candidate in the stability analysis. In the last decay, pH-based modeling and control have attracted much attention and provided significant outcomes \cite{Schaft2014,Fu2019,Dai2020,Gritli2017_2}. 
For instance, nonprehensile manipulation of a rolling robotic system based on the pH framework and a passivity approach was investigated in~\cite{Serra2019}. 
Based on the energy-shaping method, a static and a dynamic feedback control scheme were applied to stochastic pH systems in~\cite{Wassim2018}. 
New studies were carried out by inserting the impulse effects on pH systems.
Recently, a study on the stabilization problem and \( H_\infty \) control for a switched pH system with actuator saturation was published in~\cite{Ming2019}. In this last work, the stability analysis was examined by applying the dwell time method and multiple energy-based Lyapunov functions. 

To the authors’ best knowledge, no results can be found that have addressed the issues of finite-time stabilization and disturbance estimator-based robust control for switched pH systems, simultaneously.
Finite-time stability analysis is a criterion that can be used to evaluate the stability characteristics in the dynamic motion of biped robots. A finite-time control problem for stabilization of periodic orbits in an underactuated biped robot based on a feedback linearization technique and a Poincaré map was addressed in \cite{Helin2020}. However, that control system is not robust against parameter uncertainties and external disturbances: thus, due to the influence of these factors, the stability of the periodic orbits may be disturbed. Using fractional operators in designing procedure can lead to significant improvement in the transient and steady-state performance. Also, it can improve the robustness and reduce, or even eliminate, chattering phenomena~\cite{Hong2002,Ren2011}. 
Fractional calculus was used in the control of singular perturbed systems~\cite{Song2019}, solving optimal control problems~\cite{Alinezhad2017}, tracking control of legged robots~\cite{Farid2018}, fault diagnosis and classification~\cite{Yang2020}, and formation control of multi-agent systems~\cite{Jun2019}. Besides, fractional-order disturbance observer, which is derived from the integer-order type, was applied to synchronization of two uncertain fractional-order chaotic systems~\cite{Chen2017}. Despite these remarkable contributions, fractional calculus has not been applied so far to the class of hybrid systems, especially hybrid pH systems with external disturbances. Solving the problems above for hybrid pH systems using fractional calculus constitutes the framework of this research work.

\subsection{Contributions}
Motivated by the above discussions, the objective of this study is to develop a new control law based on fractional calculations for uncertain hybrid dynamics of biped robots in the pH form. The key contributions of this research work are listed below.
\begin{itemize}
    \item   The primary interest is to replace the system's total energy, the Hamiltonian, as a function of the system states.
    \item   A two-loop control system architecture is implemented, i.e., a position control loop and velocity control loop.
    \item   A fractional-order proportional-integral-derivative (PID) position controller is used in the position control loop, making the response of the closed-loop system faster. In contrast, a central fractional-order sliding mode controller is designed for the velocity loop.
    \item   Two approaches are proposed for disturbance reconstruction: the former employs a fractional calculation; the latter employs an adaptive disturbance estimator.
    \item   By accommodating the central controller with the estimated results, the robustness of the system against unknown disturbance is fulfilled.
    \item   Lyapunov stability theorem ensures the finite-time stability of the closed-loop system during both the swing phase and the impacts.
\end{itemize}

In the final part of this work, the trajectory tracking and the robustness performance of two types of biped robots under impact effects, parameter uncertainty, and external disturbance are evaluated. The results indicate the effectiveness of the proposed control algorithms.

\subsection{Outline}
The outline of this paper is organized as follows. The hybrid dynamic description of the biped robots and some preliminaries are exhibited in Section 2. The main results on the fractional-order PID position controller, the fractional-order sliding mode velocity controller, the estimators' structure, and the stability analysis of the closed-loop system are presented in Section 3. In Section 4, two numerical simulations are provided to justify the feasibility of the proposed control method and the effectiveness of the theoretical results. The conclusion is brought in Section 5 as the final part of the paper.
\section{Preliminaries and problem statement}
In this section, preliminary knowledge about hybrid dynamics of biped robots in the pH form and fractional calculus are introduced, which are employed together as the basis of the next subsections.

\subsection{Hybrid dynamic equations of the biped robots and preliminaries}
The dynamic model of biped robots consists of nonlinear differential equations for the swing phase and algebraic equations for the collision or stance phase.
The following subsections display the two phases mentioned above that are in turn collected in a hybrid model in the pH form. 

\subsubsection{Leg dynamics in swing phase}
Using Euler-Lagrange method, the governing dynamic equation of a swing leg is given by
\begin{equation}\label{eq:dyn_mod_lagrang}
M(q)\ddot{q}+C(q,\dot{q})\dot{q}+g_0(q)=B\tau,
\end{equation}
where \(q\in\mathbb{R}^n\) and \(\dot{q}\in\mathbb{R}^n\) are the state vectors of the biped robots representing the position and velocity of the joints, respectively; \(M(q)\in\mathbb{R}^{n\times n}\), \(C(q,\dot{q})\in\mathbb{R}^{n\times n}\), and \(g_0(q)\in\mathbb{R}^n\) are called the inertia matrix, the Coriolis matrix, and the vector of gravitational terms, respectively. Finally, \(B \in \mathbb{R}^{n \times m}\) is the allocation matrix for the actuation joint torques/forces \(\tau \in \mathbb{R}^m\), with $0 < m \leq n$.

\subsubsection{Constraint equations in stance phase}
Constraints on the biped robot's state appear when the following three conditions are established in a swing leg \cite{Znegui2020}:
\begin{enumerate}[i]
    \item the swing leg is ahead from the support leg in the heading direction;
    \item the movement direction of the swing leg is downward;
    \item the swing leg hits the ground surface.
\end{enumerate}

When the swing leg strikes the walking surface, discontinuities occur in the joints' position and velocity, which may yield instability of the biped robot. The impact effects on the states of the biped robot are expressed as \cite{Jessy2014}
\begin{equation}
\begin{cases}
    q^+=\Delta_{n}q^{-},\\
\dot{q}^+=\Delta_{s}\dot{q}^{-},\\
  \end{cases}
\end{equation}
where \(\Delta_{n}\in \mathbb{R}^{n \times n}\) and \(\Delta_{s} \in \mathbb{R}^{n \times n}\) are called position renaming and velocity resetting matrices, respectively. 
Besides, the apexes $+$ and $-$ denote the quantities at a time instant after and before the impact, respectively.

The following assumption is made throughout the paper.
\begin{assumption}
The impact time of the swing leg with the walking surface is instantaneous, and the support leg immediately leaves the walking surface.
\end{assumption}

\subsubsection{Hybrid model in port-Hamiltonian form}
Assuming that the contact time of a robot leg occurs at the moment \(t_{k} \in \mathbb{R}\), with \((k=0,1,2,...)\), the hybrid dynamics of the biped robots can be written as
\begin{equation}\label{eq:dyn_lagrag_hy}
\begin{cases}
    M(q)\ddot{q}+C(q,\dot{q})\dot{q}+g_0(q)=B\tau,\hspace{1.55cm} t \neq t_k,\\
    [q^{+^T},\dot{q}^{+^T}]^T = \begin{bmatrix}(\Delta_n q^-)^T & (\Delta_s \dot{q}^-)^T \end{bmatrix}^T, \hspace{0.4cm} t = t_k.
  \end{cases}
\end{equation}

Defining the momentum vector of the system as \(p=M(q)\dot{q} \in \mathbb{R}^n\), the dynamic equations~\eqref{eq:dyn_mod_lagrang} can be rewritten in the class of first-order ordinary differential equation as
\begin{equation}
\begin{cases}
    \dfrac{dq}{dt}=f(q,p)=\dfrac{\partial H(q,p)}{\partial p},\\
    \dfrac{dp}{dt}=g(q,p)+B\tau=-\dfrac{\partial H(q,p)}{\partial q}+B\tau,\\
  \end{cases}
\end{equation}
where \(f(q,p)=M^{-1} (q)p\), \(g(q,p)=\dot{M}(q)\dot{q}-C(q,\dot{q})\dot{q}-g_0(q)\), and \(H(q,p) \in \mathbb{R}\) is the Hamiltonian function of the system with the following definition
\begin{equation}\label{eq:hamil_fun}
H(q,p)=\frac{1}{2}p^TM^{-1}(q)p+V(q),
\end{equation}
where \(V(q) \in \mathbb{R}\) is the potential energy of~\eqref{eq:dyn_mod_lagrang}. 
The general pH form of~\eqref{eq:dyn_lagrag_hy} can be written as:
\begin{equation}\label{eq:ph_sys_hy}
\begin{cases}
    \dot{x}=\Omega \nabla H + G \tau,\hspace{1cm}t \neq t_k,\\
    x^+=\begin{bmatrix}(\Delta_n q^-)^T & \delta_p(x,\Delta_n,\Delta_s)^T\end{bmatrix}^T,\hspace{0.6cm}t=t_k,\\
  \end{cases}
\end{equation}
with $x=\begin{bmatrix}x_1^T & x_2^T\end{bmatrix}^T=\begin{bmatrix}q^T & p^T\end{bmatrix}^T$,
$\nabla H=\begin{bmatrix} {\left(\nabla_q H\right)}^T & {\left(\nabla_p H\right))}^T \end{bmatrix}^T=\begin{bmatrix} {\left(\dfrac{\partial H(q,p)}{\partial q}\right)}^T & {\left(\dfrac{\partial H(q,p)}{\partial p}\right)}^T \end{bmatrix}^T$, $G=\begin{bmatrix}O_{m \times n} & B^T\end{bmatrix}^T$, $\delta_p(x,\Delta_n,\Delta_s) \in \mathbb{R}^n$ the vector addressing the impact at the momenta's level, that is \(\delta_p(x,\Delta_n,\Delta_s) = p^+=M(q^+)\dot{q}^+ = M(\Delta_nq^-)\Delta_s\dot{q}^- = M(\Delta_nq^-)\Delta_s M^{-1}(q^-)p^{-}\), and
$$
\Omega=\begin{bmatrix}O_{n\times n} & I_n\\
-I_n & O_{n\times n}\end{bmatrix},
$$
where $I_n \in \mathbb{R}^{n \times n}$ and $O_{i \times j} \in \mathbb{R}^{i \times j}$ are identity and zero matrices of proper dimensions, respectively. 

\begin{theorem}
For a nonlinear system \(\dot{x}=-\Sigma \nabla H(x)\), where \(\Sigma \in \mathbb{R}^{2n \times 2n}\) is a positive definite matrix, if the energy function \(H(x)\) is chosen as
\begin{equation}\label{eq:hamilt_thm1}
H(x)=(x^TYx)^\beta
\end{equation}
where \(1<\beta<2\), and \(Y \in \mathbb{R}^{2n \times 2n}\) is a positive definite matrix, then the system is finite-time stable.
\end{theorem}

\begin{proof}
Taking time derivative of~\eqref{eq:hamilt_thm1} yields 
\begin{equation}
\begin{split}
\dot{H}(x)&=\nabla H \dot{x}\leq-\lambda_{min} (\Sigma) \nabla H^T (x)\nabla H(x)\\
&\leq-4\lambda_{min} (\Sigma) \beta^2 (x^T Yx)^{2\beta-2} (x^T Y Y^T x)\\
&\leq-4\lambda_{min} (\Sigma) \beta^2 \lambda_{min} (Y)H^\eta\\
\end{split}
\end{equation}
where \(\eta=\frac{2\beta-1}{\beta}<1\) and $\lambda_{min}(\cdot),\lambda_{max}(\cdot) \in \mathbb{R}$ represent the minimum and maximum eigenvalue of the given matrix, respectively. The above result shows the finite-time stability of the system.
\end{proof}

\begin{remark}
The value of \(\beta\) in~\eqref{eq:hamilt_thm1} determines the type of stability (i.e., finite-time or asymptotic stability). 
In particular, if  \(\beta\geq 2\) then the selected Hamiltonian function~\eqref{eq:hamilt_thm1} cannot guarantee the finite-time stability, and the obtained results decline to the traditional asymptotic stability \cite{Wang2008}.
\end{remark}

Let $0_\times \in \mathbb{R}^\times$ be the zero vector of proper dimension.
\begin{definition} [\cite{Liu2013}]
For the hybrid system~\eqref{eq:ph_sys_hy}, with \(\tau=0_m\), \(d(x,t)=0_{2n}\), and under the following condition, the system response is finite-time stable.
\begin{equation}
H(x(t_0))<b_1  \hspace{0.3cm}  \Rightarrow  \hspace{0.3cm}  H(x(t))<b_2,  \hspace{0.3cm}  \forall t \in (0,t_f]
\end{equation}
where \(0<b_1<b_2\), and \(t_f>0\).
\end{definition}

\begin{figure*}
  \includegraphics[width=\textwidth,height=4cm]{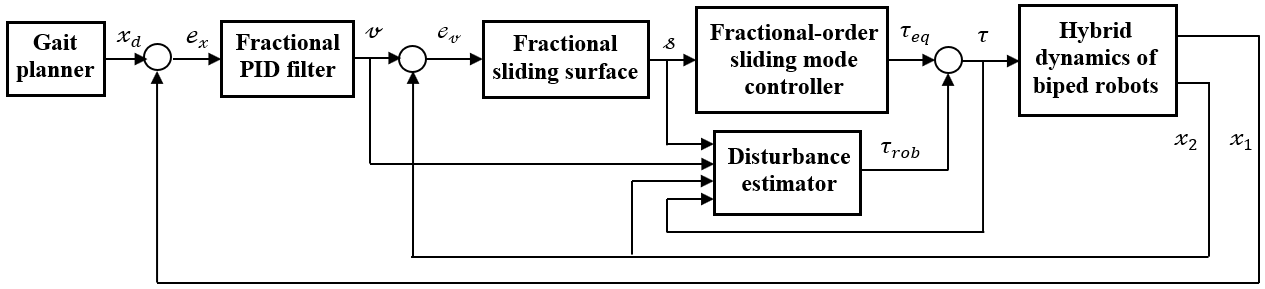}
  \caption{The devised conceptual two-loop scheme to control biped robots.}
  \label{fig:conceptual_scheme}
\end{figure*}

\subsection{Fractional calculus}
In order to fully understand the work, the fundamental definition and properties of fractional calculus are given in this section.

Among the fractional operators, which were introduced in the fractional mathematics, Caputo and Riemann–Liouville fractional operators have attracted good attentions in the field of control systems \cite{Jun2019}.
Due to great applications and well-accepted physical interpretations of Caputo’s fractional derivative, the Caputo fractional operator is employed in this work \cite{Kilbas2006}. The Caputo fractional derivative of order \(\alpha \geq 0\) is defined as
\begin{equation}
D^{\alpha}{f}(t)=\frac{1}{\Gamma(n-\alpha)}\int_{t_0}^{t}\frac{f^n(r)}{(t-r)^{\alpha-n+1}}\text{d}r,
\end{equation}
where $f(t) \in \mathbb{R}^n$ is a time-dependent function, $t_0 \in \mathbb{R}$ is the initial time of the integration, $m-1<\alpha<m$, $m\geq 1$, and $\Gamma$ is called the Gamma function with the following definition
\begin{equation}
\Gamma(z)=\int_{0}^{\infty}\iota^{z-1}e^{-\iota}\text{d}\iota.
\end{equation}

\begin{property}[\cite{Kilbas2006}]
For the optional scalars \(a_1\) and \(a_2\), the fractional orders \(\alpha_1\) and \(\alpha_2\), and two time-dependent functions \(f_1 (t)\) and \(f_2 (t)\), the following relations hold
\begin{equation}
D^{\alpha}(a_1 f_1 (t)+a_2 f_2 (t)) = a_1 D^{\alpha} f_1 (t)+a_2 D^{\alpha} f_2 (t),
\end{equation}
\begin{equation}
D^{1-\alpha}(D^{\alpha}f(t)) = \dot{f}(t),
\end{equation}
and
\begin{equation}
D^{\alpha_1}(D^{\alpha_2}f(t)) = D^{\alpha_1+\alpha_2}f(t).
\end{equation}
\end{property}

\subsection{Problem statement}
This paper's control objective is to design a fractional calculus-based estimator, an adaptive estimator for external disturbances in biped robots, and a robust centralized fractional-order sliding mode controller for rendering the states of the biped robots to the desired ones. The controller structure ensures that all signals of the closed-loop system will be finite-time stable in the swing phase and in impact time (i.e., the measured states of the system converge to the desired trajectories in a finite-time).

\section{Main theoretical results}
This section explains the design procedure of a robust fractional-order sliding mode controller based on an external estimator. The conceptual block diagram of the closed-loop control system is illustrated in Fig.~\ref{fig:conceptual_scheme}. This scheme includes a block containing the biped robot's hybrid dynamics, the gait planner block, and the control system's blocks. The gait planner produces, instead, the desired joint positions and velocities for the biped robot. The control scheme consists of two control loops: the outer and the inner loops. The outer-loop contains the position controller whose output is the desired velocity signal. At the velocity loop, besides the control of the joint velocities, the effects of parameter uncertainty and external disturbances are compensated. The devised controller structure provides better flexibility for designers to shape the biped robots' trajectory tracking response. In the following, the details of this two-loop control scheme are discussed.

Define the position tracking vector, \(e_x(t) \in \mathbb{R}^n\), and velocity tracking error vector, \(e_v(t) \in \mathbb{R}^n\), as
\begin{equation}
e_x (t)=x_1 (t)-x_{1_d}(t),
\end{equation}
\begin{equation}\label{eq:e_v}
e_v (t)=x_2 (t)-v(t),
\end{equation}
where \(x_{1_d} (t) \in \mathbb{R}^n\) is the desired position trajectory vector, and \(v(t) \in \mathbb{R}^n\) is the output of the following fractional-order proportional-integral-derivative filter
\begin{equation}\label{eq:f_pid_filter}
v(t)=K_p e_x (t)+K_d D^\alpha e_x (t)+K_i D^{-\alpha} e_x (t),
\end{equation}
where $K_p, K_d, K_i \in \mathbb{R}^{n \times n}$ are positive definite matrices, which shape the position tracking response of the controlled robotic system in the outer loop, and \(D^\alpha e_x(t)\) and  \(D^{-\alpha} e_x(t)\) represent the  fractional  derivative and fractional integral of position tracking error vector, respectively. Note that the relation~\eqref{eq:f_pid_filter} can be considered as a fractional sliding surface for the position loop. 

\begin{figure}[t]
    \centering
    \includegraphics[width=7.7cm]{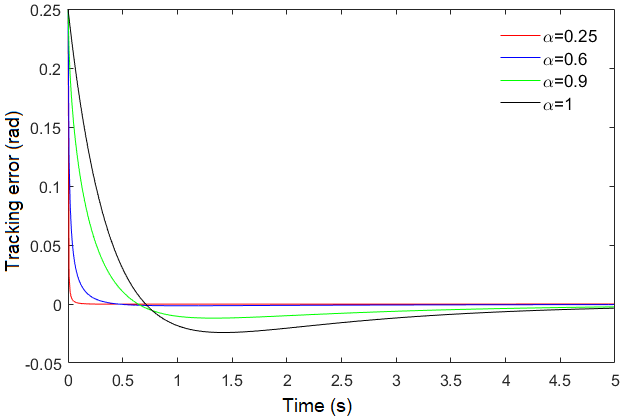}
    \caption{The convergence behavior of tracking errors for different values of fractional order $\alpha$, with $K_p=5$, $K_d=1.5$, and $K_i=2.5$.}
    \label{fig:confergence_fractional_order}
\end{figure}

\begin{remark}
Designing the fractional-order sliding surface~\eqref{eq:f_pid_filter} for the position loop has advantages with respect to its integer-order type, \(\alpha=1\), such as superior speed convergence rate, less or without overshoot/undershoot, and less steady-state tracking error. In Fig.~\ref{fig:confergence_fractional_order}, the convergence behavior of the tracking error for four values of \(\alpha\), namely \(\alpha=0.25\), \(\alpha=0.6\), \(\alpha=0.9\), and \(\alpha=1\), and with three filter gains \(K_p=5I_n\), \(K_d=1.5I_n\), and \(K_i=2.5I_n\), and for the case that the position loop is in the sliding mode, \(v(t)=0\), has exhibited. For the smaller \(\alpha\), \(\alpha=0.25\), an excellent tracking performance is observed For \(\alpha=1\), instead, the most degraded tracking performance is perceived, that is, the convergence speed decreases while the undershoot, settling time, and the steady-state tracking error increase.
\end{remark}

Taking the time derivative of~\eqref{eq:e_v}, and considering the external disturbance effect and parameter uncertainty, \(d(x,t) \in \mathbb{R}^{n}\), the dynamic equations of the velocity tracking error can be written as
\begin{equation}\label{eq:hybrid_bipedal}
\begin{cases}
    \dot{e}_v(t)=-\nabla_q H + B \tau + d(x,t)-v(t),\hspace{0.5cm}t \neq t_k\\
    e_v^+=\delta_v,\hspace{4.8cm}t=t_k\\
  \end{cases}
\end{equation}
where \(\delta_v \in \mathbb{R}^n\) is the difference between the measured and desired momenta resetting  vectors, \(\delta_v = \delta_p - \delta_{p_d}\), where \(\delta_{p_d}\) is the desired momenta resetting vector, computed from $x_{1,d} = q_d$ and $v(t)$, and where dependencies are suppressed to shorten notation. 
It is important to note that the elements of the disturbance term, \(d(x,t)\), have the dimension of velocity and an acceleration-like terms such as the time derivative of the momenta.

\subsection{Robust tracking controller design}\label{sec:cont_des}
In this section, the idea of constructing a centralized fractional-order robust sliding mode controller for the inner loop is presented. A robust control law is added to the control signal to maintain the robustness of the closed-loop system against external disturbance. The robust control law is constructed based on the estimation result of the disturbance estimator. In order to reconstruct the external disturbance, two new approaches are presented.

In the velocity loop, the following fractional non-singular terminal sliding surface is used
\begin{equation}\label{eq:sliding_surface}
s(t)=D^{-1}\left[D^\sigma e_v (t)+K_{s_1} e_v^\zeta (t)+K_{s_2} \text{sign}(e_v (t))\right],
\end{equation}
where \(K_{s_1} \in \mathbb{R}^{n \times n}\) and $K_{s_2} \in \mathbb{R}^{n \times n}$ are two positive definite matrices, $\sigma>0$, and $\zeta<1$. 
\begin{assumption}
The following Lipschitz conditions are established for the nonlinear function \(\delta_v\) and its fractional derivatives as,
\begin{equation}\label{eq:Assump2_eq1}
{\|D^{\sigma-1}\delta_v\|}^\beta \leq  \gamma_1  {\|D^{\sigma-1} e_v\|}^\beta,
\end{equation}
\begin{equation}\label{eq:Assump2_eq2}
{\|D^{-1}{(\delta_v)}^\zeta\|}^\beta \leq  \gamma_2  {\|D^{-1} {e_v}^\zeta\|}^\beta,
\end{equation}
\begin{equation}\label{eq:Assump2_eq3}
\|D^{-1}\textrm{sign}(\delta_v)\|^\beta \leq  \gamma_3  \|D^{-1} \textrm{sign}(e_v )\|^\beta,
\end{equation}
where \(\gamma_i>0\), with \(i=1,2,3\). 
\end{assumption}

The proposed centralized fractional-order robust sliding mode controller for the velocity loop has the following expression
\begin{equation}\label{eq:control_law_1}
\begin{split}
\tau &= \tau_{eq} = -{B}^{-1} [-\nabla_q H + D^{1-\sigma} ( K_{s_1}e_v^\zeta \\
&+ K_{s_2}\textrm{sign}(e_v) )-v(t)+K_{s_3} s(t)+K_{s_4} s(t)^\mu]\\
\end{split}
\end{equation}
where \(0<\mu<1\) and \(K_{s_3},K_{s_4} \in \mathbb{R}^{n \times n}\) are two positive  definite gain matrices.

In the following theorem, the result of the controller design for the pH dynamic of the introduced bipedal robot is given.
Notice that the hypothesis of neglecting the disturbance within~\eqref{eq:hybrid_bipedal} will be relaxed in the next subsections.

\begin{theorem}
Consider the hybrid dynamics of the bipedal robot in~\eqref{eq:hybrid_bipedal}, with $d(x,t) = 0_n$, and the fractional sliding surface~\eqref{eq:sliding_surface}. 
The proposed control law~\eqref{eq:control_law_1} ensures the finite-time stability of the sliding surface with the following reaching time
\begin{equation}
\begin{split}
     t_r \leq t_0 + &\frac{1}{2 \lambda_{min}(K_{s_3}) (1-\mu)}\\
     &\ln \left(\frac{\lambda_{min}(K_{s_3}) {V_s}^{\frac{1-\mu}{\beta}}+ \lambda_{min}(K_{s_4})}{\lambda_{min}(K_{s_4})}   \right ), 
\end{split}
\end{equation}
where \(V_s (\cdot)\) is the energy function of the systems~\eqref{eq:hybrid_bipedal}.
\end{theorem}

\begin{proof}
The time derivative of \(s(t)\) is
\begin{equation}\label{eq:sliding_sur_dyn}
\begin{split}
     \dot{s}(t)&= D^{-1+\sigma} (D^{2-\sigma}s(t))\\
     &=D^{-1+\sigma} \dot{e}_v(t)+K_{s_1}e_v^\zeta +K_{s_2}\text{sign}(e_v(t))\\
     &=D^{-1+\sigma} [-\nabla_q H + B \tau]
     +K_{s_1}e_v^\zeta \\
     &\quad+K_{s_2}\text{sign}(e_v(t)).
\end{split}
\end{equation}

Replacing the equivalent control law~\eqref{eq:control_law_1} yields
\begin{equation}\label{eq:s_dot}
     \dot{s}(t) =-K_{s_3}s(t) - K_{s_4}s(t)^\mu.
\end{equation}

Consider the following Lyapunov function, which is taken equal to the Hamiltonian of the overall system
\begin{equation}\label{eq:dot_V_s}
     V_s(s) = (s(t)^T Ys(t))^\beta. 
\end{equation}
Taking time derivative of \(V(s)\) yields
\begin{equation}
    \dot{V}_s(s) = 2\beta(s(t)^T Ys(t))^{\beta-1} (s(t)^T Y\dot{s}(t)).
\end{equation}
Folding~\eqref{eq:s_dot} into~\eqref{eq:dot_V_s} yields
\begin{equation}
\begin{split}
     &\dot{V}_s(s) =-2\beta(s(t)^T Ys(t))^{\beta-1} (s(t)^T Y(K_{s_3}s(t) \\ &\qquad\qquad+ K_{s_4}s(t)^\mu))\\
     & \leq -2\beta \left [\lambda_{min}(K_{s_3})V_s(s) + \lambda_{min}(K_{s_4})V_s(s)^{\frac{\beta + \mu -1}{\beta}} 
     \right],
\end{split}
\end{equation}
which indicates that the sliding surface~\eqref{eq:sliding_surface} is attractive with finite-time convergence property. 
The previous equation is rewritten as follows to calculate the reaching time
\begin{equation}
     dt \leq \frac{-dV_s(s)}{2\beta \left [\lambda_{min}(K_{s_3})V_s(s) + \lambda_{min}(K_{s_4})V_s^{\frac{\beta + \mu -1}{\beta}}(s)  \right]}.
\end{equation}
Integrating the both sides of the previous inequality, from \(t_0\) to \(t_r\), and considering that at the reaching time the value of \(V_s(t)\) is zero, \(V_s(t_r)\) and \(t_r\) can be thus obtained.
\end{proof}

\subsection{Estimator-based robustness enhancement}
In the previous section, a two-loop architecture for the closed-loop system was presented. A fractional sliding mode controller with the aim of trajectory tracking for the biped robots was proposed. In the case that the parameter uncertainties and external disturbance in the robot dynamics cannot be neglected (i.e., \(d(x,t) \neq 0_n\) in~\eqref{eq:hybrid_bipedal}), the controller structure should be adapted to eliminate or reduce their adverse effects so that the robustness of the system increases. One extensively carried out solution in the literature considers a sign function as a robust control law. Its amplitude is chosen based on the upper bound of the disturbance function \(d(x,t)\). Due to the insufficient knowledge about the maximum amplitude of $d(x,t)$, the tuning of robust control gains is not a simple work. The selection of high values for robust control gains increases the amplitude of the control effort and the system's total energy. The selection of low values for robust control gains leads to decreased system performance and increased tracking error. An efficient approach to dealing with the system dynamics' undesired signals is online control and a disturbance estimator.

Considering the disturbance estimator result, the centralized control signal \eqref{eq:control_law_1} is modified as:

\begin{equation}\label{eq:control_law_2}
\tau = \tau_{eq} + \tau_{rob}
\end{equation}
where

\begin{equation}\label{eq:control_rob}
\tau_{rob} =-{B}^{-1}\hat{d}(x,t)
\end{equation}
and \(\hat{d}(x,t)\) is the estimated value of \(d(x,t)\). 
In the following, the reconstruction of the external disturbance is explained using two approaches . 

\subsubsection{Fractional calculus-based disturbance reconstruction}\label{sec:estim_1}
The structure and the design methodology of the fractional calculus-based disturbance estimator is given in the following theorem.
\begin{theorem}
Consider the closed-loop system dynamics of a biped robot in pH form~\eqref{eq:hybrid_bipedal} and the fractional sliding surface~\eqref{eq:sliding_surface}. 
The following fractional calculus-based estimator reconstructs the value of unknown lumped disturbance function
\begin{equation}\label{eq:frac_est_dis}
\hat{d}(x,t) = \frac{1}{\rho} \left[D^{-1} \left(K_{s_3}s(t)+K_{s_4}s(t)^\mu \right) + D^{1-\sigma}s(t) \right],
\end{equation}
where \(\rho>0\) and $d(x,0) = 0_n$.
\end{theorem}

\begin{proof}
Taking the \(2-\sigma\) order fractional derivative from both sides of~\eqref{eq:sliding_surface}, substituting the closed-loop system dynamics~\eqref{eq:hybrid_bipedal}, and using the equivalent control term~\eqref{eq:control_law_1} yields
\begin{equation}\label{eq:frac_D2_dis_cal}
\begin{split}
D^{2-\sigma}&=\dot{e}_v + D^{1-\sigma}\left[K_{s_1}s^\zeta+K_{s_2}\textrm{sign}(s(t)) \right]\\
&=-\nabla_q H + B\tau + d(x,t)\\
&+D^{1-\sigma}\left[K_{s_1}s^\zeta+K_{s_2}\textrm{sign}(s(t)) \right]\\
&=d(x,t)-\hat{d}(x,t) - \left(K_{s_3}s(t)+K_{s_4}s(t)^\mu \right)\\
\end{split}
\end{equation}

The proposed estimator wants to achieve the following relationship between \(\hat{d}(x,t)\) and \(d(x,t)\) in the Laplace domain
\begin{equation}\label{eq:laplace_estimator}
\mathcal{L}[\hat{d}(x,t)]=\dfrac{1}{1+\rho \varsigma}\mathcal{L}[d(x,t)],
\end{equation}
where $\varsigma \in \mathbb{C}$ is the Laplace complex variable and $\mathcal{L}[\cdot]$ is the Laplace transform operator. Subtracting $\mathcal{L}[\hat{d}(x,t)]$ from both sides of~\eqref{eq:laplace_estimator} yields the following equation
\begin{equation}
\rho\varsigma\mathcal{L}[\hat{d}(x,t)] = \mathcal{L}[d(x,t)]-\mathcal{L}[\hat{d}(x,t)].
\end{equation}

Considering $d(x,0) = 0_n$, applying the inverse Laplace transform to the previous equation, and substituting in turn the result within~\eqref{eq:frac_D2_dis_cal} yield
\begin{equation}\label{eq:frac_D2_dis_pr}
D^{2-\sigma}=\rho\dot{\hat{\mathrm{d}}}(x,t) - \left(K_{s_3}s(t)+K_{s_4}s(t)^\mu \right).
\end{equation}

Integrating  \(D^{-1}\) in the fractional domain both sides of \eqref{eq:frac_D2_dis_pr} gives
\begin{equation}
D^{1-\sigma}=\rho\hat{\mathrm{d}}(x,t) - D^{-1}\left(K_{s_3}s(t)+K_{s_4}s(t)^\mu \right).
\end{equation}
Therefore, the reconstructed value of \(d(x,t)\) at any instant can be obtained from
\begin{equation}\label{eq:frac_dis_est}
\hat{d}(x,t) = \frac{1}{\rho} \left[D^{-1} \left(K_{s_3}s(t)+K_{s_4}s(t)^\mu \right) + D^{1-\sigma}s(t) \right].
\end{equation}

\end{proof}

\begin{remark}
The designed fractional calculus-based disturbance estimator~\eqref{eq:frac_dis_est} can be referred to as a dynamic-free estimator. Indeed, it uses only the variable of fractional sliding surface to estimate the unknown lumped disturbance.
\end{remark}

\subsubsection{Adaptive estimator-based disturbance reconstruction}\label{sec:estim_2}
In this section, the reconstruction of external disturbances using an adaptive estimator is investigated.
The proposed adaptive disturbance estimator is described by
\begin{equation}\label{eq:adaptive_dis_est}
\begin{cases}
    \hat{d}(x,t)&=\phi(x,t) + K_{e_1} x_2\\
    \dot{\phi}(x,t) &= K_{e_1} (-\nabla_q H + v(t) - \tau - \hat{d}(x,t)) \\&\quad + K_{e_2} \textrm{sign}(s(t))
  \end{cases}
\end{equation}
where $K_{e_1},K_{e_2} \in \mathbb{R}^{n \times n}$ are positive definite gain matrices.
The following assumption is considered in the following.
\begin{assumption}
The disturbance function continuously changes without any disruption over time. Besides, the norm of \(\dot{d}(t)\) is bounded, i.e., \(\|\dot{d}(t)\|\leq l_d\), with $l_d>0$.
\end{assumption}

\begin{theorem}
Consider the hybrid tracking error dynamics of a biped robot~\eqref{eq:hybrid_bipedal} under Assumption 3 and using the external disturbance estimator~\eqref{eq:adaptive_dis_est}. 
In case the fractional sliding mode controller~\eqref{eq:control_law_1} is applied and the following parameters are designed as indicated below
\begin{equation}\label{eq:param_selec}
\begin{cases}
    K_{e_1} = 0.5 K_{s_1},\\
    K_{e_2} = \vartheta K_{s_2},\\
    \lambda_{min}(K_{s_2}) \geq \dfrac{{(\kappa + 4\vartheta^2)}^2 + 4\kappa l_d +4 {l_d}^2 + 4\vartheta^2}{4\vartheta \kappa},
  \end{cases}
\end{equation}
where \(\kappa >0 \) and \(\vartheta>0\), then the finite-time convergence of the sliding variable dynamics \(\dot{s}(t)\) is guaranteed and the external disturbance \(d(x,t)\) is reconstructed in finite-time.
\end{theorem}

\begin{proof}
Let \(\tilde{d}(x,t) = d(x,t) - \hat{d}(x,t)\) be the disturbance estimation error. Taking its time derivative and using~\eqref{eq:hybrid_bipedal} give
\begin{equation}
     \dot{\tilde{d}}(t) = -K_{e_1}\tilde{d}(x,t) - K_{e_2} \text{sign}(s(t)) + \dot{d}(x,t).
\end{equation}

Applying the equivalent controller~\eqref{eq:control_law_1} to the sliding surface dynamics~\eqref{eq:sliding_sur_dyn} and considering~\eqref{eq:control_law_2}-\eqref{eq:control_rob} yield
\begin{equation}
     \dot{s}(t) = -K_{s_1}s(t) -K_{s_2} s(t)^\mu + \tilde{d}(x,t).
\end{equation}

Define the following new state vector \(\chi \in \mathbb{R}^{2n}\) as
\begin{equation}\label{eq:new_sli_dis_vec}
     \chi =\begin{bmatrix}\chi_1^T & \chi_2^T\end{bmatrix}^T = \begin{bmatrix}(s(t)^\mu )^T & \tilde{d}(x,t)^T\end{bmatrix}^T.
\end{equation}
Considering Assumption 3, the following relationship is established between \(\dot{d}(x,t)\) and \(\chi_1\)
\begin{equation}
     \dot{d}(x,t) = \|s(t)\|^{-\mu} \overline{\omega} \chi,
\end{equation}
where \(\overline{\omega} \leq l_d\). 
Taking the time derivatives of \(\chi_1\) and \(\chi_2\) in \eqref{eq:new_sli_dis_vec} and setting \(\mu=0.5\) give
\begin{equation}\label{eq:new_vec_dyn0}
\begin{cases}
    \dot{\chi}_1 = \dfrac{1}{2}\left(\|s(t)\|^{-0.5} \left(-K_{s_2}\chi_1 + \tilde{d}(x,t) \right) -K_{s_1}\chi_1\right),\\
    \dot{\chi}_2 = \|s(t)\|^{-0.5} \left(-K_{e_2} \chi_1 + \overline{\omega} \right) -K_{e_1} \chi_2.
  \end{cases}
\end{equation}

Equation~\eqref{eq:new_vec_dyn0} can be transformed into a matrix form as
\begin{equation}\label{eq:new_vec_dyn1}
\dot{\chi} = \|s(t)\|^{-0.5}A\chi + B\chi,
\end{equation}
where \(A=\begin{bmatrix}-0.5K_{s_2} & 0.5I_n\\(\overline{\omega}I_n-K_{e_2} & O_{n\times n}\end{bmatrix}\) and \(B=\begin{bmatrix}-0.5K_{s_1} & O_{n\times n}\\O_{n\times n} & -K_{e_1}\end{bmatrix}\). 

Choose the Lyapunov function as
\begin{equation}
V_e(\chi) = (\chi^T Y\chi)^\beta, 
\end{equation}
where \(Y=\begin{bmatrix}(\kappa+2\vartheta^2)I_n & -2\vartheta I_n\\-2\vartheta I_n & I_n\end{bmatrix}\). 
Differentiating \(V_e (\chi)\) with respect to time and using~\eqref{eq:new_vec_dyn1} yield
\begin{equation}\label{eq:adaptive_est_Vdot}
\begin{split}
 &\dot{V}_e(\chi) = \beta(\chi^T Y\chi)^{\beta-1} (\dot{\chi}^T Y\chi+\chi^T Y\dot{\chi})\\
 &=\beta(\chi^T Y\chi)^{\beta-1} (\|s(t)\|^{-0.5}\chi^T(A^TY + YA)\chi\\
 &+\chi^T(B^TY + YB)\chi)=\\
 &-\beta \lambda_{min}(Y^{-1}Q_1)\|s(t)\|^{-0.5}V_e(\chi) \\
 &-\beta \lambda_{min}(Y^{-1}Q_2)V_e(\chi),
\end{split}
\end{equation}
where \(Q_1=-(A^T Y+YA)\) and \(Q_2=-(B^T Y+YB)\).
In detail, these matrices have the following structure
\(Q_1=\)  
$\begin{bmatrix}
(\kappa + 4\vartheta^2)K_{s_2} + 4\vartheta(\overline\omega I_n - K_{e_2}) & N\\
K_{e_2} - \vartheta K_{s_2} - 0.5(\kappa + 4\vartheta^2)I_n-\overline{\omega}I_n & 2\vartheta
I_n\end{bmatrix}$   
and
\(Q_2=\) 
$\begin{bmatrix}
(\kappa + 4\vartheta^2)K_{s_1} & -\vartheta K_{s_1} - 2\vartheta K_{e_1} \\
-\vartheta K_{s_1} - 2\vartheta K_{e_1} & 2K_{e_1}
\end{bmatrix},$ 
where \(N = K_{e_2} - \vartheta K_{s_2} - 0.5(\kappa + 4\vartheta^2)I_n-\overline{\omega}I_n\). According to the structure of \(Q_1\) and \(Q_2\), if the parameter’s values are selected as in~\eqref{eq:param_selec}, then positive eigenvalues are obtained from \(Q_1\) and \(Q_2\). In this case, a stable response is provided for the system~\eqref{eq:new_vec_dyn1}. 

The radially unbounded feature of the Lyapunov function \(V_e(\chi)\) can be expressed as
\begin{equation}\label{eq:rad_unbound}
    \lambda_{min}(Y^\beta) \|\chi\|^{2\beta} \leq {(\chi^TY\chi)}^\beta \leq \lambda_{max}(Y^\beta) \|\chi\|^{2\beta}.
\end{equation}
From \eqref{eq:new_sli_dis_vec}, it is possible to write down the following
\begin{equation}\label{eq:new_vec_2beta}
\begin{split}
\|\chi\|^{2\beta} &= \|s(t)^\mu\|^{2\beta} + \|\tilde{d}(x,t)\|^{2\beta} \\
&\geq \|s(t)\| + \|\tilde{d}(x,t)\|^{2\beta} \geq \|s(t)\|.
\end{split}
\end{equation}
Using~\eqref{eq:rad_unbound} and~\eqref{eq:new_vec_2beta}, one can write
\begin{equation}\label{eq:sliding_bound}
\|s(t)\| \leq \|\chi\|^{2\beta} \leq {\left (\dfrac{V_ds(s)}{\lambda_{min}(Y^\beta)}   \right)}^{\dfrac{1}{\beta}}.
\end{equation}
From \eqref{eq:sliding_bound}, it follows that
\begin{equation}\label{eq:inv_sliding_bound}
    \|s(t)\|^{-0.5} \geq {\left (\dfrac{V_s(s)}{\lambda_{min}(Y^\beta)}   \right)}^{\dfrac{-1}{2\beta}}.
\end{equation}
Considering \eqref{eq:inv_sliding_bound}, equation \eqref{eq:adaptive_est_Vdot} can be rewritten as
\begin{equation}
\begin{split}
     \dot{V}_e(\chi) \leq& -\beta \lambda_{min}(Y^{-1}Q_1) {\left (\dfrac{V_s(s)}{\lambda_{min}(Y^\beta)}   \right)}^{\dfrac{-1}{2\beta}}  V_e(\chi)\\ 
     &-\beta \lambda_{min}(Y^{-1}Q_2)V_e(\chi)\\
     &\leq -\dfrac{\beta \lambda_{min}(Y^{-1}Q_1)}{(\lambda_{min}(Y^{\beta}))^{\frac{-1}{2\beta}}}{V_e}(\chi)^{\frac{\beta-0.5}{\beta}}\\
     &-\beta \lambda_{min}(Y^{-1}Q_2)V_e(\chi).
\end{split}
\end{equation}
Hence, considering Theorem 2, after the finite-time $t_r$, $\chi=0_{2n}$ is achieved. This means $s(t)=0_n$ and $\tilde{d}(x,t)=0_n$. 
Therefore, the finite-time stability is obtained for the sliding variable $s(t)$ and the exact value of $d(x,t)$ is precisely reconstructed within the time $t_r$ (i.e., $\hat{d}(x,t)=d(x,t)$).
\end{proof}

\subsection{Stability analysis in swing phase and impact time}
In subsections~\ref{sec:cont_des},~\ref{sec:estim_1}, and~\ref{sec:estim_2}, the following general form for the derivative of the Lyapunov functions was obtained
\begin{equation}\label{eq:Lya_dot_gener_form}
\dot{V}(t) \leq -aV(t) - bV^c(t), 
\end{equation}
where \(a>0\), \(b>0\), and \(0<c<1\). 
Assuming that \(t\in [t_k,t_{k+1})\), integrating~\eqref{eq:Lya_dot_gener_form} from \(t_k\) to \(t\) yieds
\begin{equation}\label{eq:Lya_fun_extention}
\begin{split}
      V^{1-c} (t) \leq& V^{1-c} (t_k ) e^{-a(1-c)(t-t_k ) }\\
      &-\frac{b}{a} [1-e^{-a(1-c)(t-t_k ) } ]. 
\end{split}
\end{equation}

For stability analysis, taking into account the hybrid nature of the system due to the impact of the swing leg, at first, the value of the sliding surface variable~\eqref{eq:sliding_surface} is calculated after the impact
\begin{equation}\label{eq:sliding_surface_plus}
\begin{split}
s^+&=D^{\alpha-1} e_v^+ +K_{s_1} D^{-1} {(e_v^+)}^\zeta + K_{s_2 } D^{-1} \text{sign}(e_v^+ )\\
&=D^{\alpha-1} \delta_v+K_{s_1} D^{-1} {(\delta_v)}^\zeta + K_{s_2 } D^{-1} \text{sign}(\delta_v).
\end{split}
\end{equation}
Using Assumption 2 and the equations~\eqref{eq:Assump2_eq1}-\eqref{eq:Assump2_eq3}, the above expression can be bounded as follows
\begin{equation}
\|s^+\| \leq \varrho_0 \left (  \|D^{\alpha-1} e_v \|+ \|D^{-1} {e_v}^\zeta \|+ \|D^{-1} \text{sign}(e_v )\|  \right),
\end{equation}
where $\varrho_0$=max($\gamma_1$,$\gamma_2\lambda_{max}(K_{s_1})$,$\gamma_3\lambda_{max}(K_{s_2})$).
Calculating the difference of the common Lyapunov function~\eqref{eq:Lya_dot_gener_form} between the post- and the pre-impact times gives
\begin{equation}\label{eq:V_diff}
\begin{split}
V&(s({t_k}^+))-V(s({t_k}^-)) = {({s^+}^TYs^+)}^\beta - {({s^-}^TYs^-)}^\beta\\
&\left({\left(\lambda_{max}(Y){\varrho_0}^2\right)}^\beta - {\left(\lambda_{min}(Y){\xi}^2\right)}^\beta\right)\\
&\left (\|D^{\alpha-1} e_v \|+ \|D^{-1} {e_v}^\zeta \|+ \|D^{-1} \text{sign}(e_v )\|  \right)\\
&\leq (\varrho-1) V(s({t_k}^-))\\
\end{split}
\end{equation}
where \(\varrho = {(\frac{\lambda_{max}(Y){\varrho_0}^2}{\lambda_{min}(Y)\xi^2})}^\beta\), and \(\xi = min(\lambda_{min}(K_{s_1},\lambda_{min}(K_{s_2})\). 
Therefore, from~\eqref{eq:V_diff}, it yields
\begin{equation}\label{eq:Lya_fun_before_after}
V(s({t_k}^+)) \leq \varrho V(s({t_k}^-)).
\end{equation}
One can design the parameters \(Y\), \(K_{s_1}\), \(K_{s_2}\), \(\gamma_1\), \(\gamma_2\), and \(\gamma_3\) so that \(\varrho>1\). 
Combining~\eqref{eq:Lya_fun_extention} and~\eqref{eq:Lya_fun_before_after} yields
\begin{equation}\label{eq:Lya_fun_extention1}
\begin{split}
&V^{1-c} (t) \leq \varrho V^{1-c} ({t_k}^-) e^{-a(1-c)(t-t_k ) }\\
&-\frac{b}{a} [1-e^{-a(1-c)(t-t_k ) } ]\\
&\leq \varrho V^{1-c} ({t_{k-1}}) e^{-a(1-c)(t-t_{k-1} ) }\\
&-\frac{b}{a} [1-e^{-a(1-c)(t-t_{k-1} ) } ]-\frac{b}{a} \varrho [1-e^{-a(1-c)(t-t_{k-1} ) } ]\\
&\leq ... \\
&\leq \varrho^{n_\sigma} V^{1-c} ({t_0}) e^{-a(1-c)(t-t_0 ) }-\frac{b}{a} [1-e^{-a(1-c)(t-t_k ) } ]\\
&-...-\frac{b}{a} \varrho^{n_\sigma} [1-e^{-a(1-c)(t-t_0) } ]\\
&\leq \varrho^{n_\sigma} V^{1-c} ({t_0}) e^{-a(1-c)n_\sigma t_N }-\\
& \frac{b}{a}  \frac{\varrho e^{-a(1-c) t_N} (e^{-a(1-c)t_N} -1) }{\varrho  e^{-a(1-c)t_N} }   [1-\varrho^{n_\sigma}  e^{-a(1-c)n_\sigma t_N } ],\
\end{split}
\end{equation}
where \(t_N = t_k-t_{k-1}\) and \(n_\sigma\) is the number of switching times. 
The bound in~\eqref{eq:Lya_fun_extention1} can be rewritten as
\begin{equation}
      V^{1-c}(t) \leq \Psi\varrho^{n_\sigma}e^{-a(1-c)n_\sigma T_N}  - \Phi,
\end{equation}
where \(\Phi = \dfrac{b}{a}  \dfrac{e^{-a(1-c)t_N} -1}{\varrho  e^{-a(1-c)t_N} -1}\) and \(\Psi = V^{1-c} (t_0 )-\varrho\Phi\). 
If \(\varrho  e^{-a(1-c)t_N} -1<0\), then the upper-bound of  \(V(t)\) will be positive
\begin{equation}
V(t) \leq {\left[\Psi\varrho^{n_\sigma} e^{a(1-c)n_\sigma T_N} - \Phi   \right]}^{\frac{1}{1-c}}=b_2.
\end{equation}

This means that the stability of the closed-loop system is preserved not only during the swing period of the leg, but also in presence of the contact.
Therefore, using Definition 1, the system is finite-time stable.

\section{Simulation results}
In this section, two case studies, namely, a two-link walker ($n=2$) and the RABBIT biped robot with five degree of freedoms (DOFs), were employed to verify the theoretical results proposed in the previous sections. For both case studies, the designed parameters for the position control loop are chosen as \(\alpha=0.75\), \(K_p=40I_n\), \(K_d=5I_n\), and \(K_i=15I_n\), while for the velocity control loop the assigned coefficients are \(\sigma=0.85\), \(\beta=1.75\), \(\zeta=0.5\), \(\mu=0.75\), \(K_{s_1}=25I_n\), \(K_{s_2}=5I_n\), \(K_{s_3}=15I_n\), \(K_{s_4}=10I_n\), and \(\rho=0.1\). The gains of the adaptive disturbance estimator are set as \(K_{e_1}=12.5I_n\) and \(K_{e_2}=7.5I_n\). The lumped unknown function \(d(x,t)\) is composed of two terms: the first term is related to the parameter uncertainties, \(d(x)\), while the second term is an external disturbance, \(d(t)\). Hence, it yields \(d(x,t)=d(x)+d(t)\). It is assumed that the uncertainty function of the two robotic systems is a ten percent deviation of the nominal value of the gradient of the Hamiltonian function \(\nabla H(x)\), i.e. \(d(x)=\pm0.1\nabla H(x)\).
The dynamics of two biped robots, the proposed controller, and the estimators were implemented in MATLAB/Simulink environment. Differential equations were solved using the Runge–Kutta algorithm with a sampling time of $10^{-3}$~s.

\begin{figure}[h!]
    \centering
    \includegraphics[width=7.7cm]{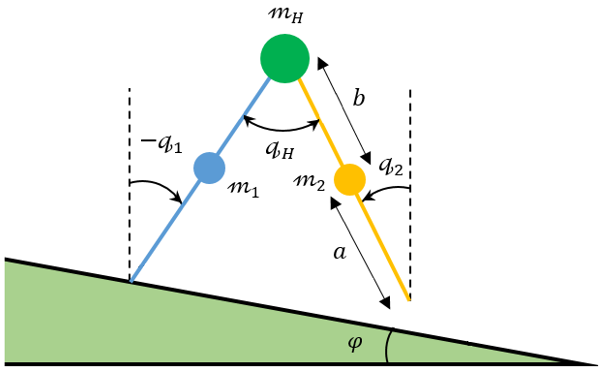}
    \caption{Schematic of a two-link walker on an inclined surface with slope $\varphi$.}
    \label{fig:two_link_walker_scheme}
\end{figure}
\begin{table}[]
\centering
\caption{Two-link walker parameters.}
\begin{tabular}{|c| c| c|} 
\hline
Symbol &	Description &	Value  \\ 
\hline
$a$   &	 Length of lower leg           &	50cm         \\
$b$   &	 Length of upper leg           &	50cm         \\
$m_1$ &	 Link1 mass	                   &    5kg          \\
$m_2$ &	 Link2 mass	                   &    5kg          \\
$m_H$ &	 Hip mass	                   &    10kg         \\
$g$   &	 Gravitational acceleration    &	9.81m/$s^2$  \\
\hline
\end{tabular}
\label{table:1}
\label{table:two_link_table}
\end{table}

\subsection{Two-link walker}
The schematic structure of the employed two-link walker is shown in Fig.~\ref{fig:two_link_walker_scheme}. 
Its geometrical parameters are instead given in Table~\ref{table:two_link_table}.
The inclined plane's angle is equal to $7.5$ degrees. The details of the dynamic equations of the employed two-link biped robot are given in Appendix A.

\begin{figure}[h!]
    \centering
    \includegraphics[width=7.4cm]{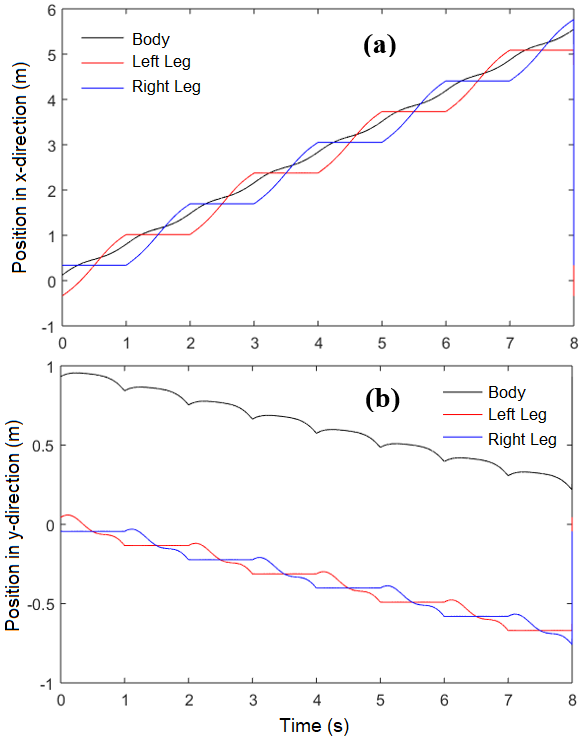}
    \caption{Cartesian positions of the two-link walker in $x$ and $y$ directions.}
    \label{fig:two_link_cartesian}
\end{figure}

\begin{figure}[h!]
    \centering
    \includegraphics[width=7.4cm]{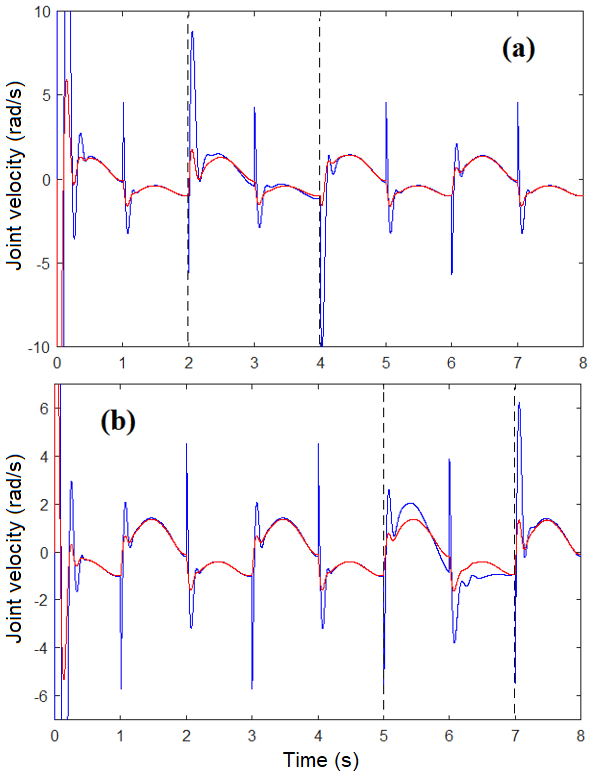}
    \caption{Desired, red line, and measured, blue line, joint velocities of the two-link biped robot without considering disturbance estimation results: (a)$1^{st}$ joint, and (b) $2^{nd}$ joint.}
    \label{fig:two_link_velocity}
\end{figure}

\begin{figure}[h!]
    \centering
    \includegraphics[width=7.4cm]{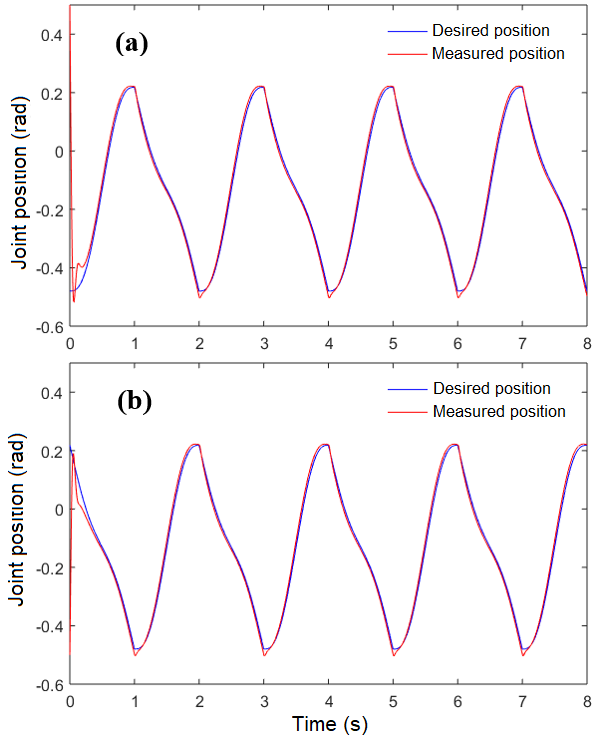}
    \caption{Desired, blue line, and measured, red line, joint angles of the two-link biped robot: (a)$1^{st}$ joint, and (b) $2^{nd}$ joint.}
    \label{fig:two_link_position}
\end{figure}

\begin{figure}[h!]
    \centering
    \includegraphics[width=6cm]{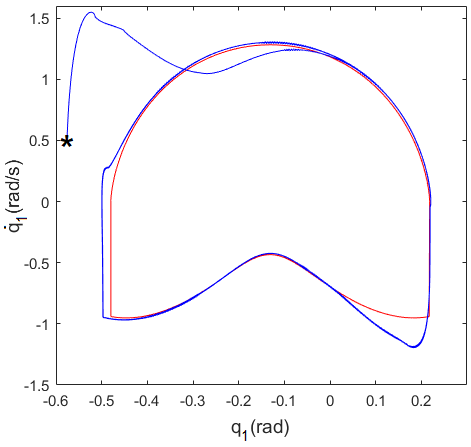}
    \caption{Phase portrait of $q_1-\dot{q}_1$ for the two-link walker: the red line is the desired phase-space, while the blue line is the measured phase-space. Marker $"*"$ indicates initial configuration at $t=0$}
    \label{fig:two_link_phase}
\end{figure}

\begin{figure}[h!]
    \centering
    \includegraphics[width=7.7cm]{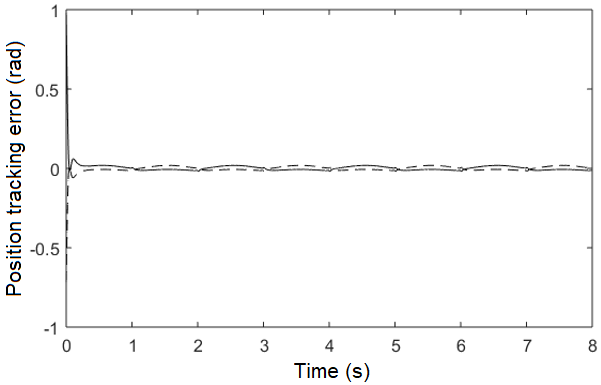}
    \caption{Position tracking errors: the black line is for joint $1$, and black dashed line is for joint $2$.}
    \label{fig:two_link_error}
\end{figure}

The gait planner~\cite{Gritli2017} produces the desired joint angles for the two-link walker. In an ideal case, the reference trajectories of the foot tips and body movement are shown in Fig.~\ref{fig:two_link_cartesian}. The variation of the angles \(q_1\) and \(q_2\) spans between \(-27.5\)~deg and \(12.5\)~deg and between \(12.5\)~deg and\(-27.5\)~deg, respectively. This depends on the swing or support phases of the legs. 
In this scenario, the swing time of each leg is set to one second.

To test robustness, the unknown external disturbances appear in the dynamic of \(\dot{q}_1\) within the time interval \(t\in [2,4]\)~s as \(d_1(t)=25\cos(2t)\), and in the dynamic of \(\dot{q}_2\) within the time interval \(t\in [5,7]\)~s as \(d_2(t)=25\sin(2.5t)\). In this case study, in order to estimate such external disturbance, the fractional calculus-based estimator is used.

\begin{figure}[h!]
    \centering
    \includegraphics[width=7.7cm]{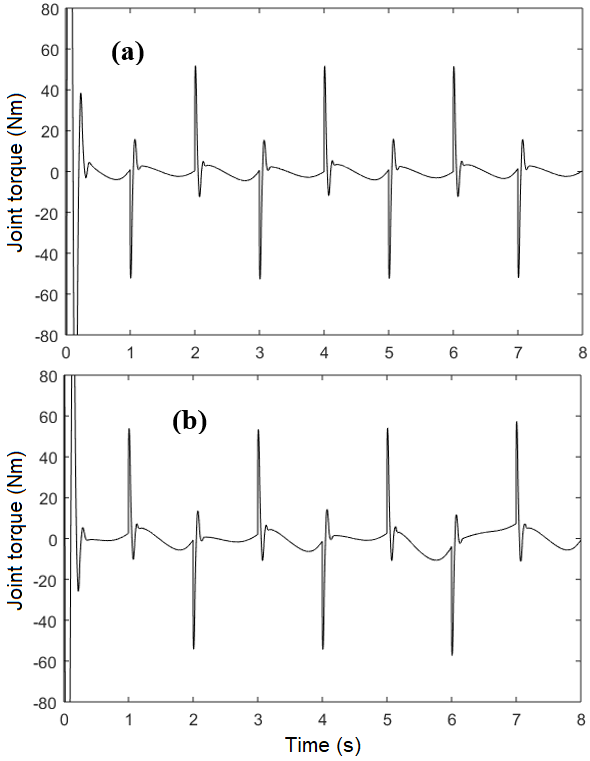}
    \caption{Control efforts of two link walker: (a) $\tau_1$, and (b) $\tau_2$.}
    \label{fig:two_link_control}
\end{figure}

Here, at first, the situation where the external disturbances' estimation results are not employed within the control structure is addressed. The effects of such external disturbances on the tracking performance of the velocity loop are shown in Fig.~\ref{fig:two_link_velocity}. In this figure, the external disturbances are applied within the two dashed lines' time interval. It can be observed that the external disturbance affects the behavior of the reference velocity, and the measured velocities cannot follow it. Therefore, the tracking performance of the system decreases and needs some modifications in the control signal. Fig.~\ref{fig:two_link_position} shows the profiles of the desired and measured angular position of the joints after adding the estimation results in the control signal. It can be observed that, now, the measured angles track the desired trajectories precisely in a finite-time when external disturbance and parameter uncertainties affect the robot dynamics.

\begin{figure}[h!]
    \centering
    \includegraphics[width=5cm]{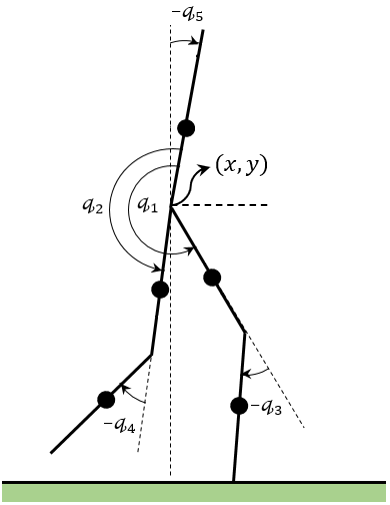}
    \caption{Schematic of the five link RABBIT biped robot.}
    \label{fig:rabbit_scheme}
\end{figure}

The phase portrait of \(q_i-\dot{q}_i\), \(i=1,2\) is plotted in Fig.~\ref{fig:two_link_phase}. It can be seen that, due to the collision of the robot leg with the ground, the dynamic behavior of the system changes. However, the system response follows the desired phase portrait in a short time. 
Fig.~\ref{fig:two_link_error} shows the time response of the angular position tracking error. It can be observed that the tracking errors decay to zero in a short time. Hence, the control algorithm has a fast convergence rate and confirms that the system is finite-time stable. Control efforts \(\tau_1\) and \(\tau_2\) are plotted in Fig.~\ref{fig:two_link_control}. At the impact times, sudden jumps appear in the control signals. The reason is that, after each impact time, the impulse effects change the values of the angular velocities sharply.  However, after a short time, the impulse effects are not observed in the control signals anymore, and the signals demonstrate continuous behaviors.

\begin{table}
\centering
\caption{RABBIT biped robot parameters.}
\begin{tabular}{|c| c| c|} 
\hline
Symbol &	Description &	Value  \\ 
\hline
$L_T$	&  Torso length           &	    63cm         \\
$L$	    &  Thigh (shank) length   & 	40cm         \\
$m_T$	&  Hip mass               &	    12kg         \\
$m_t$	&  Thigh mass             & 	6.8kg        \\
$m_s$	&  Shank mass	          &     3.2kg        \\
$I_T$	&  Hip inertia moment	  &     1.33kg$m^2$  \\
$I_t$	&  Thigh inertia moment	  &     0.47kg$m^2$  \\
$I_s$	&  Shank inertia moment	  &     0.20kg$m^2$  \\
$I_a$	&  Motor rotate inertia   &	    0.83kg$m^2$  \\
\hline
\end{tabular}
\label{table:rabbit_table}
\end{table}

\subsection{RABBIT biped robot}
The considered RABBIT biped robot is a planar mechanical system with five DOFs as llustrated in Fig.~\ref{fig:rabbit_scheme}. Its structure parameters are provided in Table~\ref{table:rabbit_table}. 
For numerical simulation, the same dynamic model of the biped robot in ~\cite{Eric2007} has been employed. 
The mathematical relationships describing the changes in the angular position and angular velocity of the joints at the impact time are given in Appendix B. 

\begin{figure}[h!]
    \centering
    \includegraphics[width=7.5cm]{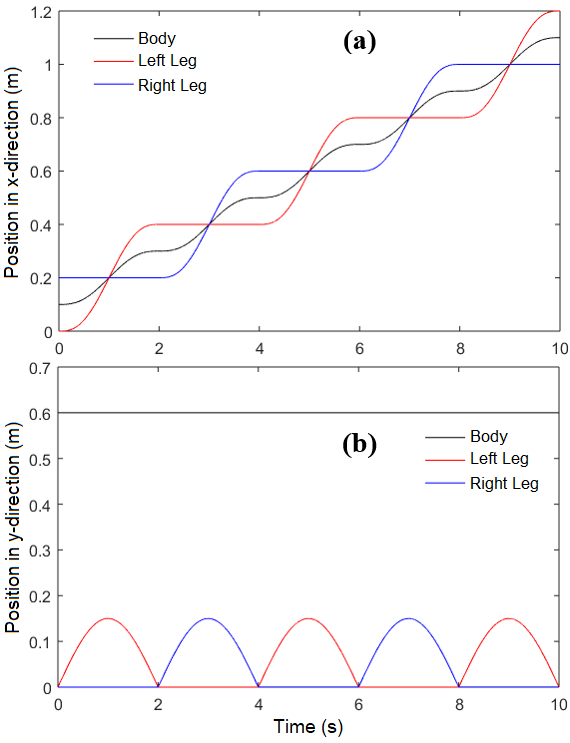}
    \caption{Cartesian positions of the RABBIT biped robot in $x$ and $y$ directions.}
    \label{fig:rabbit_cartesian}
\end{figure}

The initial standing positions of the hip, left leg, and right leg are $(0.1,0.6)$~m, $(0,0)$~m, and $(0,0.2)$~m, respectively. 
For this biped robot, at first, the Cartesian positions of the hip and tip points of the legs are designed. Afterwords, the joint positions are calculated using inverse kinematic calculations. It is assumed that, during the movement cycle, the robot's hip moves with a constant height and the tips of the legs follow the paths shown in Fig.~\ref{fig:rabbit_cartesian} in an ideal case. The swing time of each leg is set to two seconds.

To test robustness, it is assumed that the external disturbance affects only the dynamics of \(\dot{q}_1\) and \(\dot{q}_4\) as \(d_1(t)=30\cos(1.5t))\) for \(t\in[2, 5]\)~s and \(d_4(t)=25\sin(1.5t))\) for \(t\in[6, 9]\)~s.

\begin{figure}[h!]
    \centering
    \includegraphics[width=6.8cm]{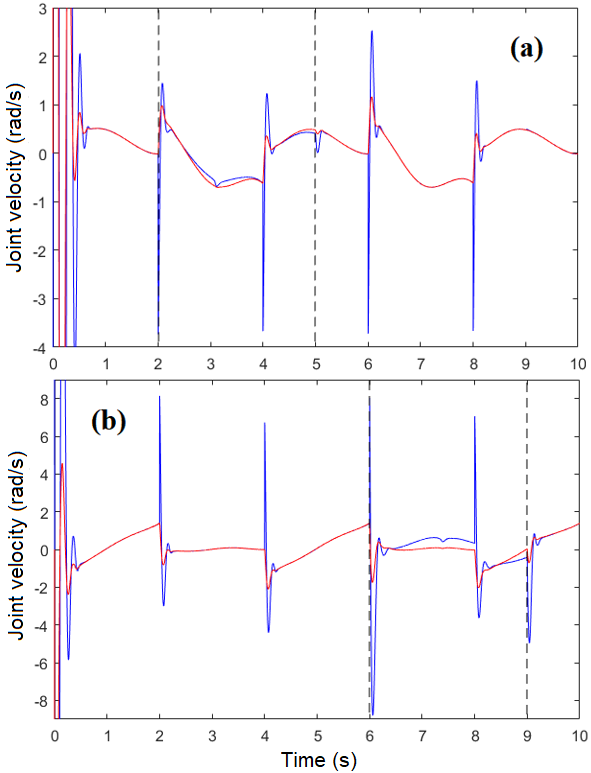}
    \caption{Desired, red line, and measured, blue line, joint velocities of the RABBIT biped robot without considering disturbance estimation results: (a)$1^{st}$ joint, and (b) $4^{th}$ joint.}
    \label{fig:rabbit_velocity}
\end{figure}

\begin{figure}[h!]
    \centering
    \includegraphics[width=6.8cm]{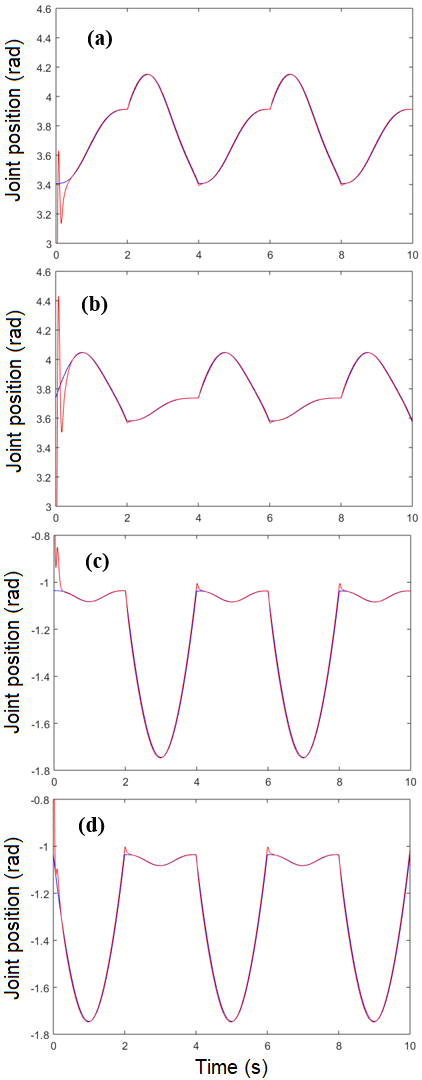}
    \caption{Desired, blue line, and measured, red line, joint angles of the RABBIT biped robot: (a) $1^{st}$ joint, (b) $2^{nd}$ joint, (c) $3^{rd}$ joint, and (d) $4^{th}$ joint.}
    \label{fig:rabbit_position}
\end{figure}

\begin{figure}[h!]
    \centering
    \includegraphics[width=6cm]{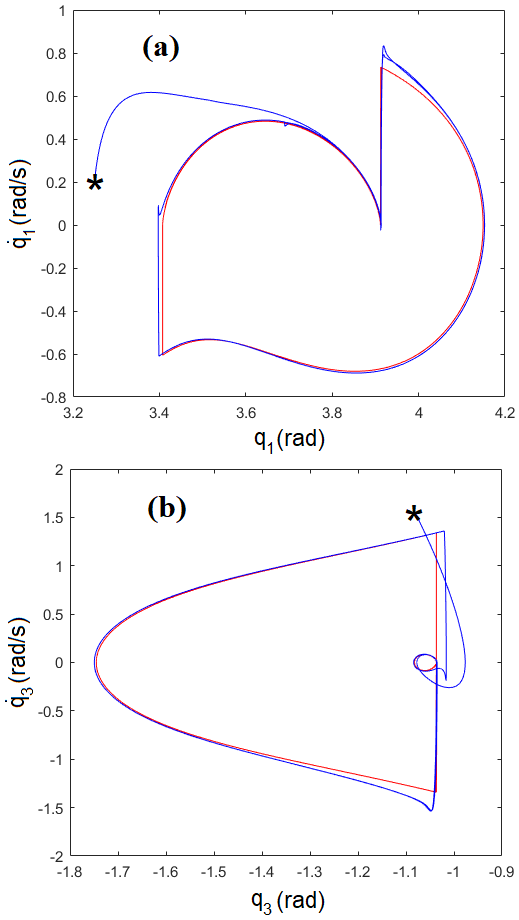}
    \caption{Phase portraits of the desired, red line, and measured, blue line, states of the RABBIT biped robot. Markers $"*"$ indicate initial configuration at $t=0$}
    \label{fig:rabbit_phase}
\end{figure}

\begin{figure}[h!]
    \centering
    \includegraphics[width=7.7cm]{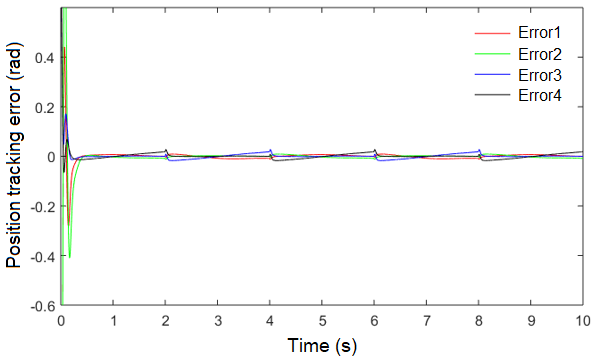}
    \caption{Position tracking errors.}
    \label{fig:rabbit_error}
\end{figure}

\begin{figure}[h!]
    \centering
    \includegraphics[width=6.8cm]{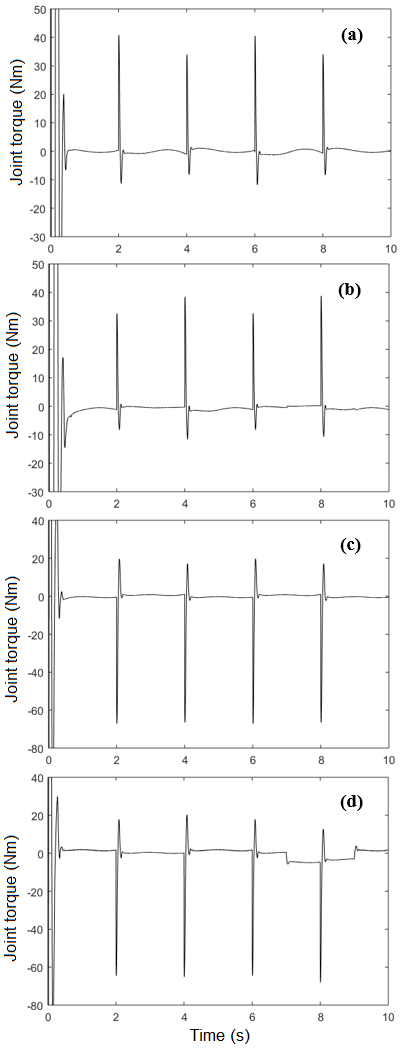}
    \caption{Control efforts of RABBIT biped robot: (a) $\tau_1$, (b) $\tau_2$, (c) $\tau_3$, and (d) $\tau_4$.}
    \label{fig:rabbit_control}
\end{figure}

At first, to evaluate the robustness of the proposed two-loop control scheme, only the equivalent control term, \(\tau_{eq}\), is applied to the RABBIT robot dynamics.
The velocity tracking results are depicted in Fig.~\ref{fig:rabbit_velocity}. From such a figure, it is evident that the velocity tracking performance falls over the time that the external disturbance affects the dynamics of the RABBIT biped robot. In such conditions, the compensation of the disturbance effect is indeed demanded. 

Here, the adaptive disturbance estimator is instead used to reconstruct the disturbance. The results in controlling the RABBIT biped robot under the proposed methodology are displayed from Fig.~\ref{fig:rabbit_position} to Fig.~\ref{fig:rabbit_control}. 
In detail, the time responses of the robot states and the desired joint trajectories are plotted in Fig.~\ref{fig:rabbit_position}. 
The results are representative of the good convergence of the measured joint position states to their desired trajectories. Fig.~\ref{fig:rabbit_phase} illustrates the phase diagrams between the states \(q_i\) and \(\dot{q}_i\), \(i=1,3\). It also shows that the system responses are adapted to impact conditions and follow the desired phase portraits. Since the states \(q_2\) and \(q_4\) behave similarly to \(q_1\) and \(q_3\), their phase spaces are not displayed. Since the gait planner produces stable motion for the RABBIT biped robot, and the robot goes on stable limit-cycles, the dynamic stability of the robot is preserved~\cite{Jessy2014}. Fig.~\ref{fig:rabbit_error} shows the behavior of the position tracking errors under fractional-order sliding mode controller. The position tracking errors tend to zero as quickly as expected. 
Fig.~\ref{fig:rabbit_control} presents the applied control signals \(\tau_1\) to \(\tau_4\). 
It can be observed that, due to impacts, spike-like signals appear, but they are repelled by the control system quickly. These results verify the theoretical prediction and validate the effectiveness of the proposed control scheme.

To further test the robustness of the control system and better evaluate the proposed estimators' performance, the amplitude of the external disturbances is also increased $1.5$ times. The position tracking results obtained in the time interval $[0.5,10]$~s are compared in terms of the root-mean-square error (RMSE). 
The RMSE obtained by using the adaptive estimator is $0.0106$ and $0.0139$ for the \(1^{st}\) and the \(4^{th}\) joint, respectively. The RMSEs obtained applying the fractional calculus-based estimator are $0.0461$ and $0.0395$ for the \(1^{st}\) and the \(4^{th}\) joint, respectively.
Since the adaptive estimator uses the robot dynamics, better performance is achieved than the fractional calculus-based estimator.

\section{Conclusion}
The problem of designing a finite-time robust controller for the hybrid dynamics of a biped robotic system subject to impact, parameter uncertainty, and external disturbance was investigated. 
Considering the fractional sliding surface's excellent convergence properties, a fractional-order PID position controller was used in the position loop.  A fractional-order sliding mode controller was instead applied in the velocity loop. 
Two disturbance reconstruction techniques were proposed to deal with unknown uncertainty and disturbance function. The former, using fractional calculus properties, can reconstruct the unknown lumped disturbance. The latter is an adaptive disturbance estimator. 
The stability of the closed-loop hybrid port-Hamiltonian system was proved using Lyapunov theory. 
The achieved results obtained using two types of biped robots bolster the goodness of the proposed control methodology. 
Besides, the proposed finite-time control strategies foresee an effective approach to solve the finite-time control problems for general switched nonlinear port-Hamiltonian systems. 

Future plans want to extend this work to the switched port-Hamiltonian systems and only for the case that the system's position variables are measurable. It will also be assumed that there is a time-delay between the control action and the controlled system's switching time. That is, the asynchronous switching controller design problem for port-Hamiltonian systems will be investigated.
To further prove the proposed methodology in practice, the designed controller can be implemented to the real biped robots with several degrees of freedom under different walking/running conditions.

\section*{Appendix A}  
The details of two link walker dynamic matrices are as follows \cite{Eric2007}:

\(M\) = 
$\begin{bmatrix}
M_{11} & M_{12} \\
M_{21} & M_{22}
\end{bmatrix}$

\(M_{11} = (m_H+m_2)l^2+m_1 b^2 \)

\(M_{12} = -m_2 la \cos(q_1-q_2)\)

\(M_{21} = -m_2 la \cos(q_1-q_2)\)

\(M_{22} = m_2 a^2\)

\(C\) = 
$\begin{bmatrix}
C_{11} & C_{12} \\
C_{21} & C_{22}
\end{bmatrix}$

\(C_{11} = 0 \)

\(C_{12} = -m_2 la\dot{q_2} \sin(q_1-q_2)\)

\(C_{21} = m_2 la\dot{q_1} \sin(q_1-q_2)\)

\(C_{22} = 0\)

\(G\) = 
$\begin{bmatrix}
-((m_H+m_2)l+m_1 b)g \sin(q_1) \\
m_2 ag \sin(q2)
\end{bmatrix}$

where \(l=a+b\).

The renaming and resetting matrices \(\Delta_n\) and \(\Delta_s\) have the following forms:

\(\Delta_n\) = 
$\begin{bmatrix}
0 & 1 \\
1 & 0 \\
\end{bmatrix}$

\(\Delta_s = {Q_+}^{-1} Q_-\) ,

where:

\(Q_+\) = 
$\begin{bmatrix}
{Q_+}_{11} & {Q_+}_{12} \\
{Q_+}_{21} & {Q_+}_{22}
\end{bmatrix}$

\({Q_+}_{11} = m_1 a^2 -m_1 la \cos(q_1-q_2)) \)

\({Q_+}_{12} = -m_1 la \cos(q_1-q_2))+(m_H+m_1)l^2+m_2 b^2\)

\({Q_+}_{21} = m_1 a^2\)

\({Q_+}_{22} = -m_1 la \cos(q_1-q_2)) \)

\(Q_-\) = 
$\begin{bmatrix}
{Q_-}_{11} & {Q_-}_{12} \\
{Q_-}_{21} & {Q_-}_{22}
\end{bmatrix}$

\({Q_-}_{11} = (m_H l^2+(m_1+m_2)lb)\cos(q_1-q_2)) \)

\({Q_-}_{12} = -m_2 ab\)

\({Q_-}_{21} = -m_1 ab\)

\({Q_-}_{22} = 0\)

From Fig. 4, the impact time of the swing leg can be recognized from the following equation:

\(P(q)=P_1 (q) + P_\phi(q)- P_2(q)=0\)

where:

\(P_1 (q)=l \cos(-q_1))\)

\(P_2 (q)=l \cos(q_2))\)

\(P_\phi (q)=L \tan(\phi), \hspace{0.2cm} L=l \sin(-q_1))+l \sin(q_2))\)

\section*{Appendix B} 
For the RABBIT biped robot, $\Delta_n$ has the following structure:

\(\Delta_n\) = 
$\begin{bmatrix}
0 & 1 & 0 & 0 & 0 \\
1 & 0 & 0 & 0 & 0 \\
0 & 0 & 0 & 1 & 0 \\
0 & 0 & 1 & 0 & 0 \\
0 & 0 & 0 & 0 & 1 
\end{bmatrix}$

And, the relation between after and before joint velocities, \(\Delta_s\), is calculated by the following equation \cite{Eric2007}:

\(\Delta_s = I-M^{-1}J^T {[J M^{-1} J^T]}^{-1}J\) 

where \(J\) is the Jacobian matrix of the swing leg with respect to the \((x,y)\) position in Fig.~\ref{fig:rabbit_scheme}.

\section*{Acknowledgment}
The research leading to these results has been supported by both the PRINBOT project (in the frame of the PRIN 2017 research program, grant number 20172HHNK5\_002) and the WELDON project (in the frame of Programme STAR, financially supported by UniNA and Compagnia di San Paolo). The authors are solely responsible for its content.

\bibliography{mybibfile}

\end{document}